\documentclass[letterpaper]{article} 
\usepackage[]{aaai25}  
\usepackage{times}  
\usepackage{helvet}  
\usepackage{courier}  
\usepackage[hyphens]{url}  
\usepackage{graphicx} 
\usepackage{natbib}  
\usepackage{caption} 
\usepackage{algorithm}
\usepackage{algorithmic}
\usepackage{newfloat}
\usepackage{listings}
\DeclareCaptionStyle{ruled}{labelfont=normalfont,labelsep=colon,strut=off} 
\floatstyle{ruled}
\newfloat{listing}{tb}{lst}{}
\floatname{listing}{Listing}

\usepackage{amsfonts}
\usepackage{amsmath}
\usepackage{lipsum} 
\usepackage{multirow}

\title{Representing Neural Network Layers as Linear Operations \\ via Koopman Operator Theory}
\author {
    Nishant Suresh Aswani\textsuperscript{\rm 1,2},
    Saif Eddin Jabari\textsuperscript{\rm 1,2},
    Muhammad Shafique\textsuperscript{\rm 1,2}
}
\affiliations {
    \textsuperscript{\rm 1} New York University Tandon, Brooklyn, USA \\
    \textsuperscript{\rm 2} New York University Abu Dhabi, Abu Dhabi, UAE\\
    nishantaswani@nyu.edu, sej7@nyu.edu, muhammad.shafique@nyu.edu
}

\begin{document}

\maketitle

\begin{abstract}
The strong performance of simple neural networks is often attributed to their nonlinear activations. However, a linear view of neural networks makes understanding and controlling networks much more approachable. We draw from a dynamical systems view of neural networks, offering a fresh perspective by using Koopman operator theory and its connections with dynamic mode decomposition (DMD). Together, they offer a framework for linearizing dynamical systems by embedding the system into an appropriate observable space. By reframing a neural network as a dynamical system, we demonstrate that we can replace the nonlinear layer in a pretrained multi-layer perceptron (MLP) with a finite-dimensional linear operator. In addition, we analyze the eigenvalues of DMD and the right singular vectors of SVD, to present evidence that time-delayed coordinates provide a straightforward and highly effective observable space for Koopman theory to linearize a network layer. Consequently, we replace layers of an MLP trained on the Yin-Yang dataset with predictions from a DMD model, achieving a mdoel accuracy of up to \(97.3\%\), compared to the original \(98.4\%\). In addition, we replace layers in an MLP trained on the MNIST dataset, achieving up to \(95.8\%\), compared to the original \(97.2\%\) on the test set. 
\end{abstract}

%

\begin{figure*}[t!]
\centering
\includegraphics[width=\textwidth]{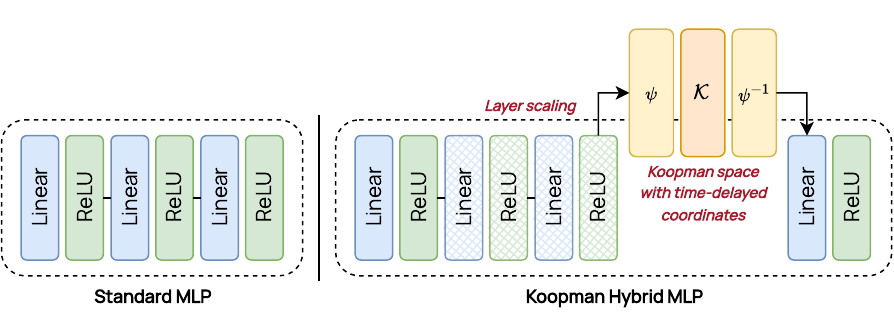}
\caption{Comparing our Koopman hybrid approach to a standard model. (Left) A typical MLP with compositions of Linear (blue) + ReLU (green) layers; (Right) Our proposed layer linearization approach, which includes scaling the layer (hatched boxes) and ``lifting" the activations into the ``Koopman space" via delay coordinates embedding (yellow and orange), consequently replacing the original layer to obtain a Koopman hybrid model.}
\label{fig:hero}
\end{figure*}

\section*{Introduction}

Trained neural networks arrive at a decision by applying a series of transformations to their inputs. Each layer plays a specific, albeit often difficult to interpret, role in achieving this goal. At every step, there is a nonlinear mapping to an abstract space, often resulting in a change of dimensionality. Under this view, neural networks are among the simplest instances of a discrete dynamical system \cite{e_proposal_2017}. Given the nature of our optimization tools, we do not arrive at systems with explicit formulae describing the dynamics. Nonetheless, we do find ourselves in a data rich regime, still allowing us to use powerful tools to study complex dynamical system. In this work, we use Koopman operator theory, a well-established approach dynamical systems to represent nonlinear systems with a linear operator, making it particularly effective due to its data-driven nature.

While an emerging body of work applies Koopman theory to study neural networks, research at this intersection has largely focused on treating the \textit{optimization} procedure as a dynamical system, targeting speedup in training and better understanding weight initialization \cite{dogra_optimizing_2020, mohr_applications_2021}. Our work differs in how we draw the link between Koopman theory and neural networks, instead focusing on the \textit{layers of the network} as a composition of dynamical systems. Outside of Koopman theory, this reframing is relatively well researched; a wealth of literature has described neural networks as dynamical systems, with a significant effort directed towards developing a framework \cite{e_proposal_2017,thorpe_deep_2022} for residual networks, resulting in new architectures \cite {lu_beyond_2020} and training approaches \cite{chang_multi-level_2018}. Although our work draws from the same shift in perspective, we direct our attention towards linearizing individual nonlinear layers in the neural network, investigating the impact on model performance, and working towards a more interpretable understanding of network layers.

The work \cite{sugishita_extraction_2024} most closely aligned with ours investigates a similar perspective, validating the use of Koopman theory in linearizing neural networks. However, they focus on linearizing the entirety of the intermediate layers, limiting the architectural choices available for their analysis. Moreover, they primarily rely on a monomial embedding as the choice of observable function. On the other, we successfully demonstrate the use of delay coordinates embedding---a straightforward yet powerful procedure---as an observable function, which is a key ingredient when implementing Koopman theory in practice. 


\section*{Background}

\subsection{Koopman Operator Theory}
In the classical perspective, a nonlinear dynamical system which evolves the system state \(\mathbf{x} \in \mathbb{R}^n\) from a discrete step \(k\) to \(k+1\) is described as:
\begin{equation}
\mathbf{x}_{k+1} = \mathbf{F}(\mathbf{x}_k),
\end{equation}
where \(\mathbf{F}: \mathbb{R}^n \rightarrow \mathbb{R}^n\) is the nonlinear map. Typically, such systems are analyzed by linear approximation near fixed points, along with other well developed tools in dynamical systems theory.

The Koopman operator theory \cite{koopman_hamiltonian_1931, koopman_dynamical_1932} provides an alternative approach to linearizing a nonlinear system by studying its evolution in the observable space, where \(\psi : \mathbb{R}^n \rightarrow \mathbb{C}^m\) is a function which acts on a system state to generate an observable. Koopman theory proposes a linear, infinite-dimensional operator \(\mathcal{K}\) which advances the observable of our system state \(\psi(\mathbf{x}_k)\) from one discrete step to the next. The system is described as: 
\begin{equation}
\psi(\mathbf{x}_{k+1}) = \mathcal{K}\psi(\mathbf{x}_k),
\end{equation}
where \(\mathbf{x}_k\) is first ``lifted" into the observable space and then advanced by $\mathcal{K}$, producing \(\psi(\mathbf{x}_{k+1})\), an evolved state in the observable space.

In practice, Koopman analysis requires a finite-dimensional approximation of the operator, which is obtained by restricting the set of observables to a ``Koopman-invariant subspace", such that \(\psi\) and \(\mathcal{K\psi}\) lie in the same subspace \cite{brunton_koopman_2016}. Looking towards the eigenvalue problem of a linear operator, \(\mathcal{K}\Phi= \Lambda \Phi\), where \({\Phi}\) is a set of eigenfunctions and \({\Lambda}\) is a diagonal matrix of corresponding eigenvalues, we identify eigenfunctions as a suitable candidate for observable functions. Upon action by the Koopman operator, the eigenfunctions remain in the original subspace; hence, they are ``Koopman invariant." Then, setting our observables (\(\psi\)) in the eigenfunction basis as \(\psi = \xi \Phi \), where \(\xi\) is a collection of coefficients that allow us to linearly combine our eigenfunctions (\(\phi\)), we can reinterpret the system. In fact, we can obtain the observables advanced $k$ number of steps with the equation
\begin{equation}
\psi(\mathbf{x}_{k}) = \mathcal{K}^k\xi\Phi(\mathbf{x}_0) = \Phi(\mathbf{x}_0)\Lambda^k\xi.
\label{equation:koopman_advance}
\end{equation} 

\subsection{Data-Driven Koopman Analysis}

Assuming we have access to \(k+1\) snapshots of data from steps \(0\) to \(k\), and our observables lift the state to dimension \(m\), we can build a data matrix \(\mathbf{D} \in \mathbb{C}^{m \times k}\) of observables. Here, \(\mathbf{D}\) omits the first observable. Each column \(\mathbf{D}_i \in \mathbb{C}^m\) is a state in the observable space which can be factorized as \(\Phi(\mathbf{x}_0) \Lambda^i \xi \), where only the number of actions \(i\) applied by \(\Lambda\) varies between columns. Then, the entire data matrix can also be factorized as 
\begin{equation}
\mathbf{D} = \Phi \, \text{diag(}{\xi}\text{)} \, [\mathbf{I} \; \Lambda \cdots \Lambda^{k}] = \Phi \, \text{diag($\xi$)} \, \mathbf{M}.
\label{equation:factorization}
\end{equation} 

This formulation can be recast to replace \(\mathbf{M}\), a Vandermonde matrix of eigenvalues, by looking at its relationship with companion matrices. Vandermonde matrices diagonalize companion matrices in the manner \(\mathbf{C} = \mathbf{M}^{-1} \text{diag($\xi$)} \mathbf{M}\) \cite{krake_dynamic_2022}. Notably, the companion matrix has a square structure with a row-shifted diagonal and a nonzero last column: 
\begin{equation}
\mathbf{C} = \begin{bmatrix}
0 & 0 & \cdots & 0 & c_0 \\
1 & 0 & \cdots & 0 & c_1 \\
0 & 1 & \cdots & 0 & c_2 \\
\vdots & \vdots & \ddots & \vdots & \vdots \\
0 & 0 & \cdots & 1 & c_{n-1}
\end{bmatrix}.
\end{equation}
Introducing the diagonalization to Equation \ref{equation:factorization}, results in
\begin{equation}
\mathbf{D} = \Phi \, \text{diag($\xi$)} \, \mathbf{M} = \Phi \, \mathbf{M} \, \mathbf{C} = \mathbf{D}' \mathbf{C},
\label{equation:new_factorization}
\end{equation}
where the final expression substitutes \(\Phi \, \mathbf{M}\) with \(\mathbf{D}'\). With the exception of the final column, we know that each column \(\mathbf{D}'_i\) is identical to \(\mathbf{D}_{i+1}\) because we know the action of \(\mathbf{C}\). The final column \(\mathbf{D}'_k\), however, is a linear combination of all the previous snapshots. We can formulate the last expression in Equation \ref{equation:new_factorization} as a minimization problem to solve for \(\mathbf{C}\). 

Overall, this discussion leads us to a framework for obtaining the eigenfunctions and eigenvalues of the Koopman operator directly from the data itself. Reformulating the framework as an algorithm results in an early version of the core dynamic mode decomposition (DMD) method. The more recent and standard, core DMD algorithm alleviates computational issues in the described approach. Our brief description largely follows the review by Schmid \shortcite{schmid_dynamic_2022}, which provides a detailed look at developments in the DMD algorithm. Nevertheless, our exposition of the original algorithm emphasizes the use of observed measurements to analyze a nonlinear dynamical system. We build our work atop this data-driven foundation.


\begin{figure}[t!]
\centering
\includegraphics[width=0.95\columnwidth]{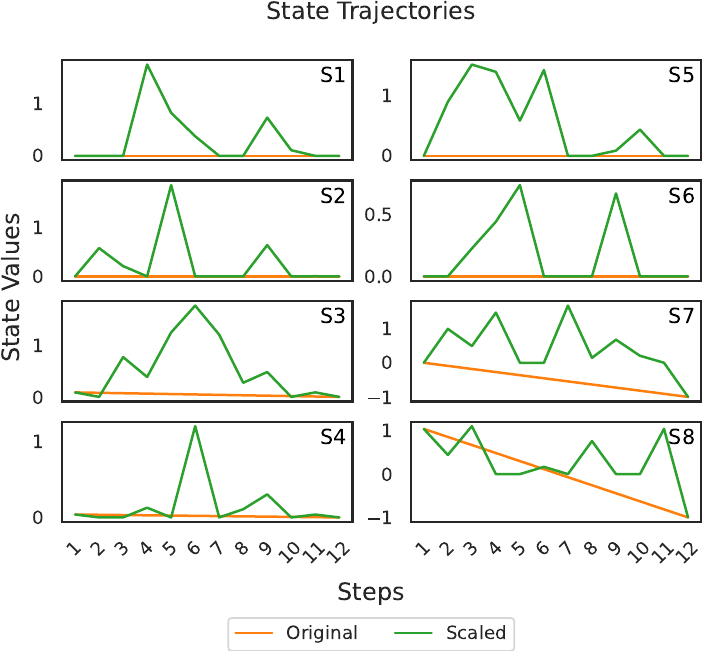}
\caption{A sample trajectory with 8 states (S1-8) from an original (orange) and scaled (greened) \(8 \times 6\) Linear + ReLU layer in an MLP trained on the Yin-Yang dataset. States S7,8 are augmented with \(-1\) on the output to allow for a trajectory of system states with uniform dimensionality.}
\label{fig:yinyang_states}
\end{figure}

\begin{figure*}[t!]
\centering
\includegraphics[width=\textwidth]{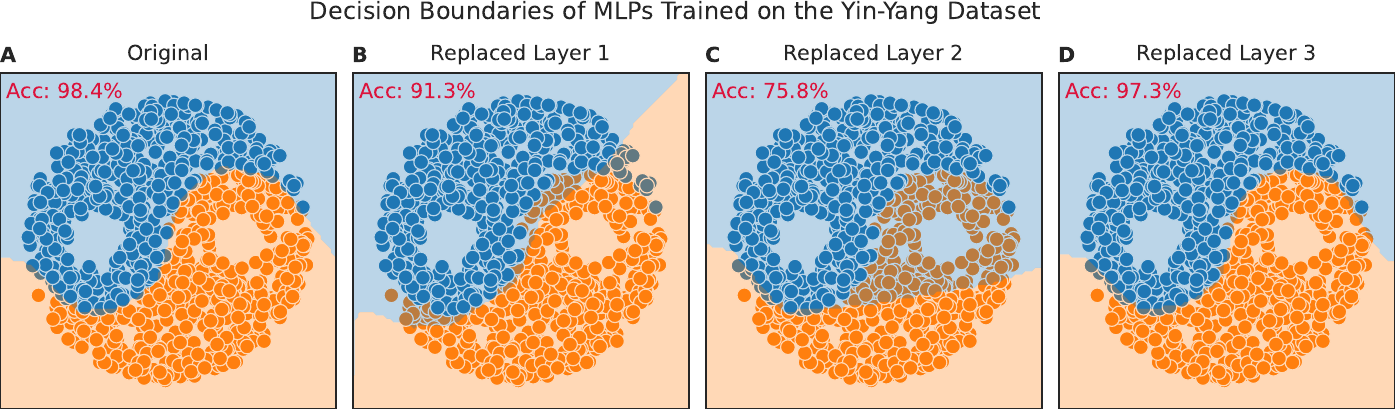}
\caption{Decision boundaries of the original MLP, and its hybrid variants, trained on the Yin-Yang dataset. We test the models on \(1000\) samples of the dataset. (A) The original model achieves an accuracy of \(98.4\%\). (B) The MLP where the first hidden layer of size \(8 \times 6\) is replaced with a DMD model, achieves an accuracy of \(91.3\%\). (C) MLP with second \(6 \times 4\) hidden layer replaced, achieves an accuracy of \(75.8\%\). (D) MLP with final \(4 \times 3\) hidden layer replaced, achieves an accuracy of \(97.3\%\).}
\label{fig:yinyang_boundary}
\end{figure*}

\section*{Koopman Theory for Neural Networks}
Our work adapts the Koopman framework to study trained neural networks by treating them as sequential compositions of nonlinear transformations. Each layer is an individual nonlinear mapping that advances a set of states forward to the next step.

\subsection{Reformulating Layers} Specifically, we consider trained multi-layer perceptrons (MLPs) as \( \mathcal{F} = \mathbf{F}_{\ell-1} \circ \cdots \circ \mathbf{F}_{1} \circ  \mathbf{F}_0 \), where each layer \( \mathbf{F}_i: \mathbb{R}^{d_i} \rightarrow \mathbb{R}^{d_{i+1}} \) for \(i \in [0,\ell-1] \cap \mathbb{Z}\) consists of a linear operation and a ReLU. We treat the inputs, activations, and outputs as system states denoted by \(\mathbf{x}_i \in \mathbb{R}^{d_i}\), where \(d_0 = n\) represents the input dimension and \(d_{\ell}\) represents the output dimension. The dimensionality of our inputs and activations is directly equal to the number of states in our system. Our framework lifts \(\mathbf{x}_i\) to \(\psi(\mathbf{x}_i)\) and replaces \(\mathbf{F}_i\) with a linear operator \(\mathcal{K}_i\). Assuming we apply this approach to the final layer \(\mathbf{F}_{\ell-1}\), we would represent model inference as
\begin{equation}
\mathbf{x}_l =  \psi^{-1} \circ \mathcal{K}_{\ell-1} \psi \circ \cdots \circ \mathbf{F}_1 \circ \mathbf{F}_0(\mathbf{x}_0),
\end{equation}
where \(\psi^{-1}\) is the inverse function which returns the observable to the original space.

\subsection{Dimensionality of the System State} 
Our goal is to predict the output of \(\mathbf{F}_1(\cdot) \in \mathbb{R}^{d_2}\). But for neural networks, the dimensionality of our system state may vary between layers; it is not guaranteed that \(d_2 = d_1 \). In our work, we admit two variants of fully-connected layers, ones that do not affect dimensionality (\(d_2 = d_1\)) and decoder layers that reduce dimensionality (\(d_2 < d_1\)). In the latter case, we augment the output of \(\mathbf{F}_1(\cdot)\) by appending \(d_1 - d_2\) negative integers to the system state. Given that our MLPs use the ReLU activation, except for this augmentation, the system never encounters negative integers.

\subsection{Layer Scaling}
Standard literature presents Koopman theory's applications to dynamical systems with inherent time steps, such as fluid flow and financial engineering \cite{brunton_koopman_2016, schmid_dynamic_2022}. Canonically, there are no time steps in model inference: each layer acts on its inputs only once to produce activations for the subsequent layer. However, DMD's performance improves with more samples \cite{schmid_dynamic_2022}. For this purpose, we apply \textit{layer scaling}, which inserts and trains additional layers in an otherwise frozen neural network, without significantly affecting the original model's performance. 

Consider \( \mathcal{F} = \mathbf{F}_{1} \circ \mathbf{F}_0 \), a trained and frozen two-layer MLP, where we are interested in replacing the final layer \(\mathbf{F}_{1}\). We scale the network by inserting an additional set of layers \(\mathcal{G} = \mathbf{G}_{k-1} \circ \cdots \circ \mathbf{G}_{1} \circ  \mathbf{G}_0 \), where \( \mathbf{G}_j: \mathbb{R}^{d_1} \rightarrow \mathbb{R}^{d_1} \) for \(j \in [0,k-1] \cap \mathbb{Z} \). Then, we train \(\mathcal{G}\) alone to minimize the Huber loss between the outputs of \(\mathbf{F}_{1} \circ \mathbf{G}_{k-1}\) and \(\mathbf{F}_{1} \circ \mathbf{F}_0\). Consequently, we obtain a scaled network \(\widetilde{\mathcal{F}} = \mathbf{F}_{1} \circ \mathcal{G} \circ \mathbf{F}_0 \). We note that, because of how we train \(\mathcal{G}\), ablating \(\mathcal{G}\) from \(\widetilde{\mathcal{F}}\) precisely returns \(\mathcal{F}\). Introducing \(\mathcal{G}\) allows us to collect a size \(k + 2\) set of activations, of uniform dimensionality \(d_1\), to build a dataset \(\mathbf{D} \in \mathbb{R}^{d_1 \times (k+2)}\), represented as
\begin{equation}
\mathbf{D} = 
\begin{bmatrix}
\mathbf{F}_0 & \mathbf{G}_0 & \mathbf{G}_1 & \cdots & \mathbf{G}_{k-1} & \mathbf{F}_1 \\
\end{bmatrix}.
\label{equation:dataset}
\end{equation}

\subsection{Delay Coordinates as Observables} So far, we have eluded a discussion of selecting an observable function. For neural networks, the number of states, equivalent to the number of rows in \(\mathbf{D}\), corresponds to the input dimension of the layer we replace. Our analysis benefits from selecting an observable function which results in a higher-dimensional set of states. To achieve this, delay coordinates embedding, or Hankelization, is a popular choice, even when complete measurement data is available \cite{kamb_time-delay_2020}. Delay coordinates embedding lifts the state by augmenting it with past history, resulting in a Hankel matrix 
\begin{equation}
\mathcal{H}_h\mathbf{D} = 
\begin{bmatrix}
\mathbf{F}_0 & \cdots & \mathbf{G}_{k-h}   \\
\mathbf{G}_0 & \cdots & \mathbf{G}_{k-h+1} \\
\vdots       & \ddots & \vdots             \\
\mathbf{G}_{h+1} & \cdots & \mathbf{F}_{1} \\
\end{bmatrix},
\end{equation}
where \(\mathcal{H}_h\mathbf{D} \in \mathbb{R}^{hd_1 \times (k-h)}\). Each row vector is a subseries of the original timeseries. Hankelization is a simple method with a single hyperparameter, the delay parameter \(h\), which determines the length of the subseries. Finally, we apply DMD to \(\mathcal{H}_h\mathbf{D}\), resulting in a fitted DMD model. When provided with \(h+1\) steps of a system state, the fitted DMD model provides any \(k\) number of future states. In our case, we limit \(k=1\), to obtain a single step, which replaces the activation of \(\mathbf{F}_1\), effectively replacing the layer.

Figure \ref{fig:hero} provides an overview of our approach. In summary, our approach scales a trained network, builds a trajectory of system states, Hankelizes the trajectory to train a DMD model, and uses the fitted DMD model's outputs in place of the activations from the hidden layer, essentially hybridizing the neural network with the aid of Koopman theory. We refer to the networks as \textit{Koopman hybrid models}.


\def\perfA{$56.48\%$ }
\def\perfB{$83.24\%$ }
\def\perfE{$63.53\%$ }
\def\perfF{$90.21\%$ }
\begin{figure*}[t!]
\centering
\includegraphics[width=0.95\textwidth]{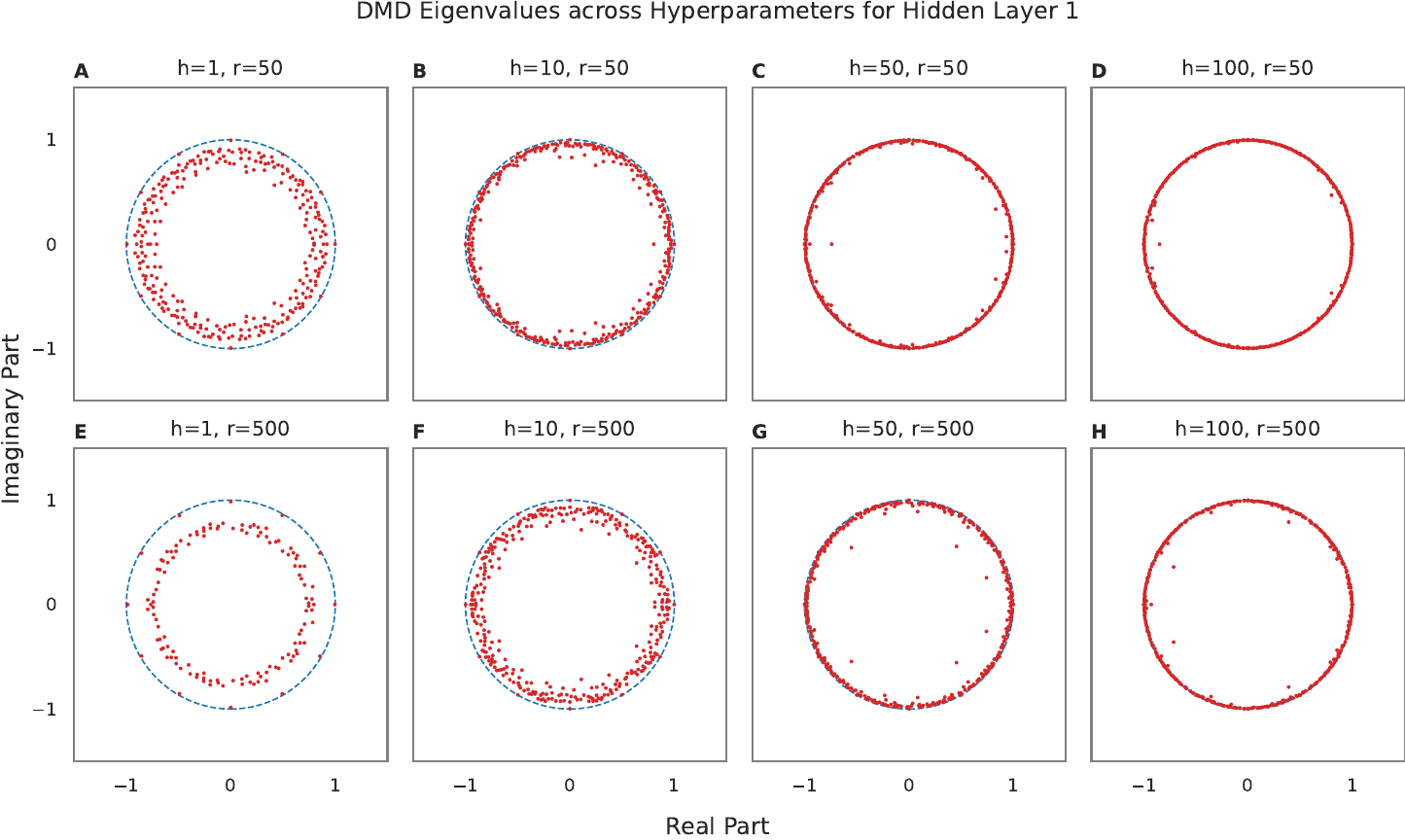}
\caption{Eigenvalue plots from varying DMD hyperparameters when replacing the first hidden layer in the MNIST network. (A-D) \(r = 50\) with \(h \in (1, 10, 50, 100)\), where (A) produces a model with \perfA accuracy and (B) \perfB accuracy; (E-H) \(r = 500\) with \(h \in \{1, 10, 50, 100\}\) where (E) produces a model with \perfE accuracy and (F) \perfF accuracy.}
\label{fig:eigenvalues}
\end{figure*}

\section*{Hybridizing a Yin-Yang Network}
\def\ogYinYangEpochs{$5000$ }
\def\ogYinYangAccuracy{$98.4\%$}
\def\ogYinYangLR{$5\mathrm{e}{-3}$}

\def\YinYangScaledLayers{$10$ }
\def\YinYangScaledEpochs{$200$ }

In this section, we train an MLP on a two-dimensional classification task, then replace its layers to visualize the effects on it's decision boundary.

\subsection{Experiment Details}
\subsubsection{Dataset.} We begin with an MLP trained on the Yin-Yang dataset \cite{kriener_yin-yang_2022}, a classification task, originally with three categories: ``Yin", ``Yang", and ``Dot", from which we exclude the latter for a simpler binary task. Each point in the adapted, two-dimensional dataset lies within one of the two categories in the ``Yin-Yang” symbol.

\subsubsection{Network.} The MLP consist of an input layer with 2 features, followed by 3 hidden layers with a collective \([ 8, 6, 4, 3]\) configuration, each using a ReLU activation, and an output layer with 2 neurons. We train the classifier on a single NVIDIA RTX 3080 using PyTorch for \ogYinYangEpochs epochs to an accuracy of \ogYinYangAccuracy, using the default Adam with decoupled weight decay (AdamW) optimizer and a learning rate of \ogYinYangLR. 

\subsubsection{Scaling and Embedding.} Each time, before layer replacement, we insert \YinYangScaledLayers additional Linear+ReLU layers and train for \YinYangScaledEpochs epochs to scale the network (see Appendix for discussion on hyperparameters). We generate a trajectory by collecting the scaled network's activations as it runs inference. 

Figure \ref{fig:yinyang_states} provides an example of a trajectory generated by a layer in a trained network and its scaled variant, where we note that the start and end states align between both networks. In addition, the final values of states S7,8 (second column) are an augmented value set to \(-1\). Here, we are scaling an \(8 \times 6\) Linear + ReLU layer, hence the augmentation affects the last \(2\) states. The process generates a matrix \(D \in \mathbb{R} ^ {8 \times 12}\), where the number of rows is determined by the number of states and the number of columns is determined by the number of scaling layers \(+\,2\). Repeating this process allows us to take advantage of more data, in turn producing a better fit DMD model. For this layer, if we use \(r\) trajectories, we obtain the data matrix \(\mathbf{D} = [D_0 \; D_1 \cdots D_{r-1}] \in \mathbb{R} ^ {8 \times 12r}\).

\subsection{Decision Boundaries of the Yin-Yang Network}
Figure \ref{fig:yinyang_boundary} illustrates the decision boundaries for the original network and three hybrid networks, each with a different hidden layer replaced by its corresponding DMD model. For this experiment, we fix the delay parameter \(h=\YinYangScaledLayers\) and the number of trajectories \(r=1000\). The boundaries reflect that our approach preserves the network's performance to varying degrees, depending on the layer replaced. After scaling the network, but prior to replacement, the changes in the decision boundaries and accuracies compared to the original model are negligible, indicating that the differences in performance must be attributed to the DMD routine, possibly due to inadequate hyperparameters. However, the maximum delay parameter is tethered to the number of steps we can provide to the DMD model, limiting our exploration. Given that we scale with \YinYangScaledLayers layers, we must set \(h=\YinYangScaledLayers\). We further explore these hyperparameters in the next section.

We hypothesize that the decline in performance indicates the complexity of the transformation undertaken by the layer we replace. Figure \ref{fig:yinyang_boundary} shows that replacing the penultimate layer has a minimal effect on the decision boundary, suggesting that it is responsible for the simplest transformation in the network's decision-making process, whereas hidden layer 2 is responsible for the most complex.

\def\perfLOneHOne{$66.66\%$ }
\def\perfLOneHTen{$92.04\%$ }
\def\perfLThreeHOne{$28.22\%$ }
\def\perfLThreeHTen{$95.80\%$ }
\begin{figure*}[t!]
\centering
\includegraphics[width=1.03\columnwidth]{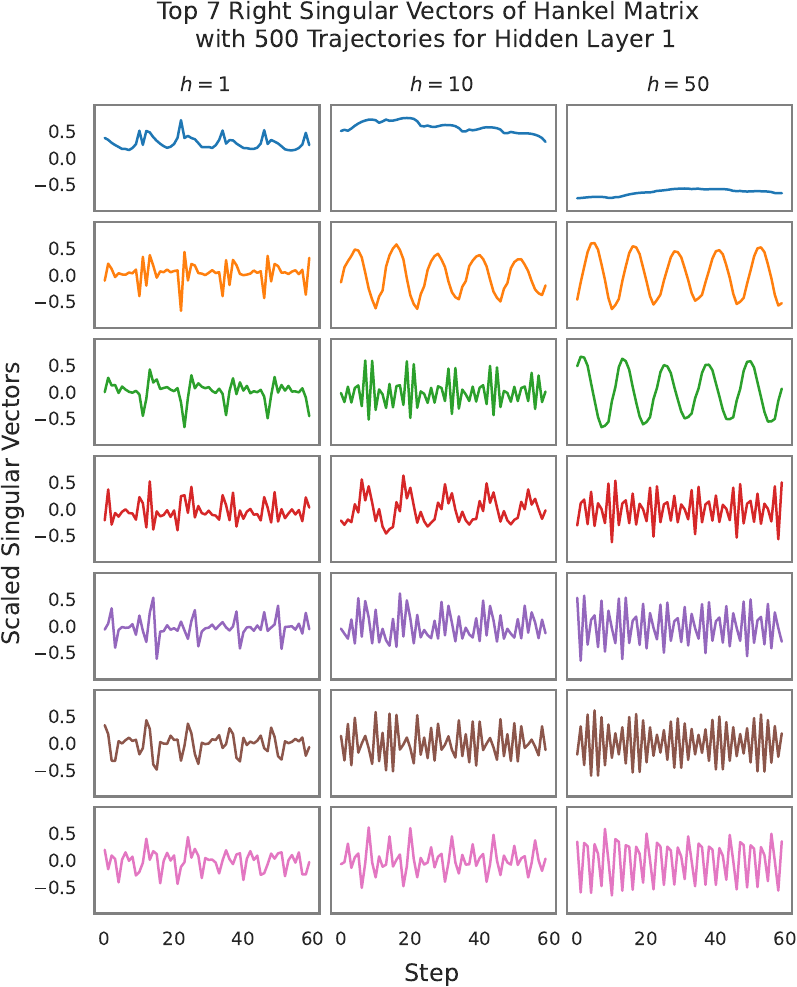}
\hfill
\includegraphics[width=1.03\columnwidth]{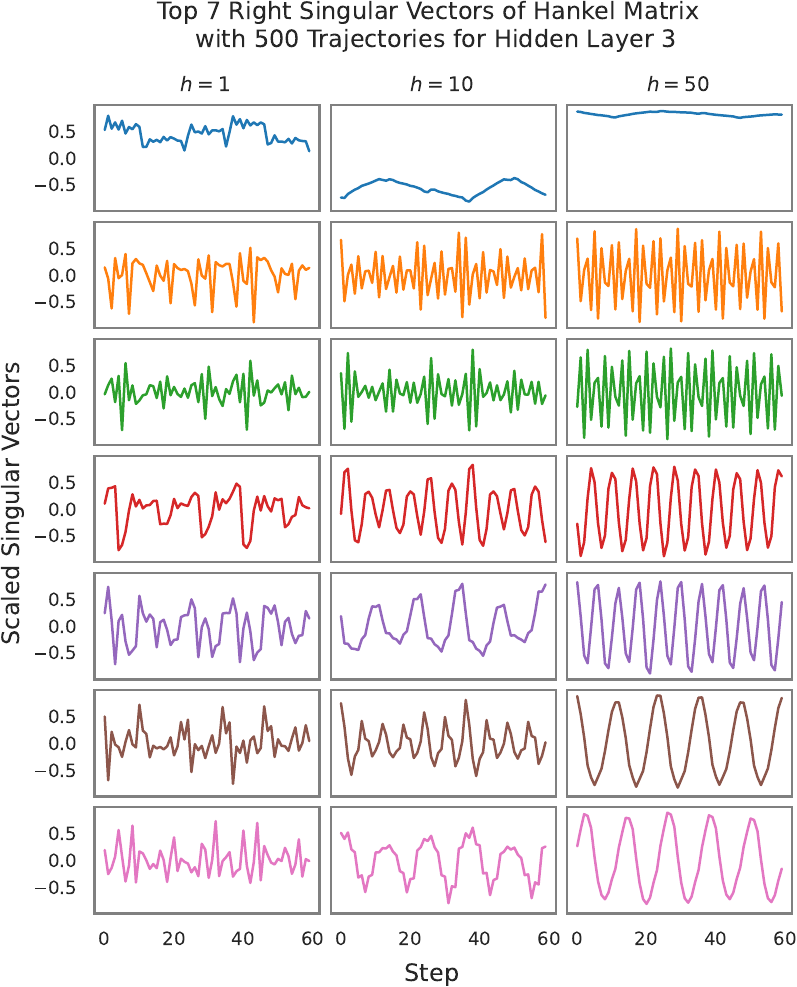}
\caption{Right singular vector (RSV) plots of the Hankel matrix with various delay parameters. (Left) the Hankel matrix is generated from the first hidden layer with \(h \in (1, 10, 50)\); \(h=1\) achieves \perfLOneHOne and \(h=10\) achieves \perfLOneHTen on the test set. (Right) the Hankel matrix is generated for the final hidden layer with \(h \in (1, 10, 50)\); \(h=1\) achieves \perfLThreeHOne and \(h=10\) achieves \perfLThreeHTen on the test set.}
\label{fig:top_rsv}
\end{figure*}

\section*{Hybridizing an MNIST Network}
To further evaluate our approach, we extend our analysis to an MLP trained on the MNIST dataset and present additional experiments to explore DMD hyperparameters.

\def\ogAccuracy{$97.20\%$ }
\def\ogEpochs{$30$ }
\def\ogLR{$1\mathrm{e}{-2}$}
\def\numTopRSV{$5$ }
\def\listDelay{\{1, 10, 10^2, 10^3\}}

\def\scaledLayers{$10$ }
\def\scaledTotalLayers{$12$ }
\def\scaledEpochs{$200$ }
\def\scaledLR{$4\mathrm{e}{-3}$}
\def\scaledBetas{$[0.85, 0.93]$ }

\def\dmdAccuracy{$95.3\%$ }

\subsection{Experiment Details}

\subsubsection{Dataset and Network.} We work with an MLP trained on MNIST digits \cite{lecun_gradient-based_1998} with a \([784, 256, 32, 16, 10]\) configuration, with the input and three hidden layers using ReLU activation. We train the classifier on a single NVIDIA RTX 3080 using PyTorch for \ogEpochs epochs to an accuracy of \ogAccuracy on the test set, using AdamW \((\beta_1=0.9, \beta_2=0.999)\) and a learning rate of \ogLR.

\subsubsection{Scaling and Embedding.} We scale with \scaledLayers Linear+ReLU layers. With the original layers frozen, the new layers are trained on the MNIST training set using the AdamW optimizer (see Appendix for a discussion on training and hyperparameters). As before, we build trajectories by collecting the network's activations as it runs inference. For example, if we analyze the \(32 \times 16\) Linear + ReLU layer, the augmentation affects the last \(16\) states and the process generates a matrix \(D \in \mathbb{R} ^ {32 \times 12}\). If we repeat the process for \(r\) samples, we arrive at a data matrix \(\mathbf{D} = [D_0 \; D_1 \cdots D_{r-1}] \in \mathbb{R} ^ {32 \times 12r}\). For this network, we scale and replace all three hidden layers.

\subsection{Exploring the Eigenvalues from Dynamic Mode Decomposition}
When Hankelizing and applying DMD to a trajectory matrix, we must select the delay parameter \(h\) and the number of trajectories \(r\). Hence, we explore the implications of these hyperparameters on the DMD routine. Figure \ref{fig:eigenvalues} shows the eigenvalues plotted alongside a unit circle for a few combinations of \(h, r\) for the first hidden layer of the network. We note that the eigenvalue plots for the two remaining hidden layers, provided in the appendix, follow a similar pattern.

Figure \ref{fig:eigenvalues} illustrates that increasing the delay parameter (left to right) increases the number of eigenvalues and eigenmodes (not visualized), implying that there are more ``building blocks" to work with. In addition, the eigenvalues tend to be pulled further out, towards the unit circle, suggesting more temporal patterns that do not decay over time. Comparing the top and bottom rows, we see that increasing the number of trajectories affects the size of the eigenvalues. In the first two columns, there are fewer eigenvalues resting on the unit circle when DMD is provided with more trajectories (E-F), compared to when provided with fewer trajectories (A-B). This may also suggest that DMD finds more decaying temporal patterns when exposed to a greater number of trajectories. Intuitively, this is sound, as it would be easier to identify more spurious patterns with a larger sample size.

It is tempting to associate improved model performance with the empirical observation of eigenvalues approaching the unit circle. In Figure \ref{fig:eigenvalues}, the settings from panel E result in an accuracy of \perfE, while those from panel F yield an accuracy of \perfF. A similar improvement appears between panel A to panel B, with accuracies of \perfA and \perfB, respectively. In both rows, the eigenvalues move outwards. But, comparing panels B and E, the correlation does not necessarily hold. We hypothesize that, while increasing the delay parameter helps the DMD model identify better ``building blocks", it must be accompanied by an increase in trajectories to avoid reliance on spurious patterns. 

\begin{figure*}[t!]
\centering
\includegraphics[width=0.95\textwidth]{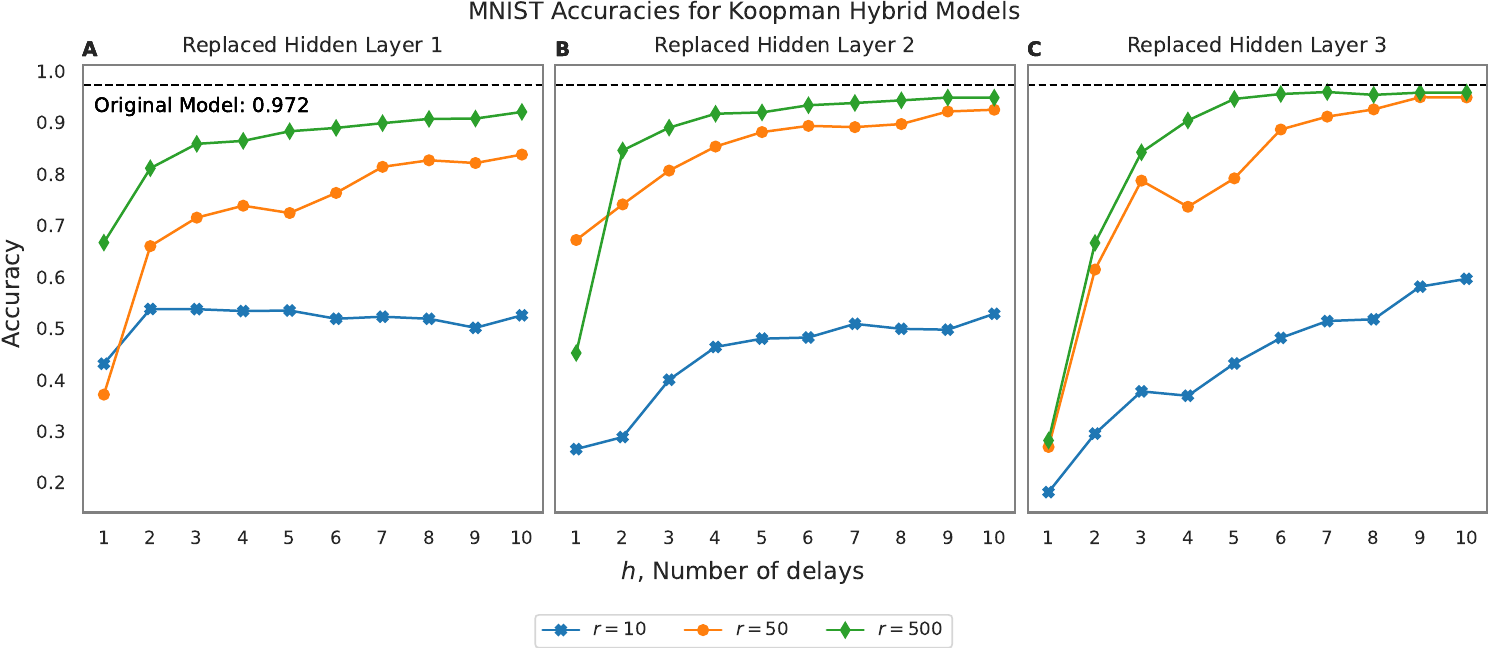}
\caption{Accuracies on the MNIST test set for various DMD hyperparameters across Koopman hybrid models. The original model achieves \ogAccuracy accuracy. (A) Accuracies after replacing hidden layer 1, where the best performing combination ($h=10, r=500$) achieves \(92.04\%\); (B) Replaces hidden layer 2, where ($h=10, r=500$) achieves \(94.81\%\); (C) Replaces hidden layer 3, where ($h=10, r=500$) achieves \(95.80\%\).}
\label{fig:mnist_accuracy}
\end{figure*}

\subsection{Exploring the Singular Vectors from Singular Value Decomposition}

Next, we explore delay embedding on the trajectory matrices, testing multiple options for the delay parameter \(h\). Here, we fix \(r = 500\). In Figure \ref{fig:top_rsv}, we compute the singular value decomposition (SVD), \(\mathcal{H}_h \mathbf{D} = \mathbf{U\Sigma V}^T \), and plot the right singular vectors \(\mathbf{V}\). In systems with a single state, both the row and column vectors of the Hankel matrix contain time sub-series, allowing one to analyze either \(\mathbf{U}\) or \(\mathbf{V}\) for temporal patterns. In our case, because the system has multiple states, the left singular vectors do not exclusively represent temporal patterns. Instead, they are a spatio-temporal mix, making them a poor candidate for inspection. Hence, we plot \(\mathbf{V}\) of the trajectory matrices generated for the first (left) and last (right) hidden layers. 

For both layers, when \(h=1\) (indicating no Hankelization), the plotted vectors are noisy, complex objects. However, Hankelizing the matrix linearizes the system in the Koopman space, as evidenced by the simpler, more uniform sinusoidal curves that emerge. Interestingly, in the first hidden layer (Figure \ref{fig:top_rsv} left), the dominant singular vector has the lowest frequency, whereas in the final hidden layer (Figure \ref{fig:top_rsv} right), the dominant singular vectors have higher frequencies. In either case, given the improvement in performance, Hankelizing the trajectory matrix is a successful strategy to linearize our dynamics.

\subsection{Network Performance After Layer Replacement}
Our goal is to replicate the activations of a layer, while maintaining the model's performance. To that end, we insert the learned eigenmodes, eigenvalues, and coefficients (see Equation \ref{equation:koopman_advance}) from DMD in place of the layer of interest and run inference on the MNIST test dataset. While previous discussions have reinforced the selection of a high delay parameter, we are limited to a maximum \(h=10\) for replacement experiments as described earlier.

Figure \ref{fig:mnist_accuracy} presents, for a combinations of parameters, the accuracies of all three hybrid MNIST classifiers, achieving up to \(95.80\%\) accuracy (replacing hidden layer 3), compared to the original \ogAccuracy. Here, the best performing combination of available hyperparameters is \(h=10, r=500\), regardless of the layer replaced. While increasing the number of trajectories clearly improves the model, the rate of improvement plateaus, as evidenced by the small difference in performance between \(r=50\) and \(r=500\), compared to \(r=10\) and \(r=50\). Supporting previous discussion, there is a positive trend in accuracy when increasing \(h\), demonstrating that delay coordinates embedding is a viable observable function for linearizing neural network layers. Even for \(r=10\), increasing the delay parameter significantly improves the hybrid model's performance. Increasing either parameter increases the computational cost to fit the DMD model, so it is advisable to select the fewest trajectories with the smallest delay that still produces adequate results. Further, in line with results from Figure \ref{fig:yinyang_boundary}, we find varying levels of success in hybridizing the model depending on the layer. As before, replacing the final hidden layer is most successful, potentially highlighting a quality of how neural networks process data.

\section*{Conclusion}
Under the lens of of dynamical systems, we demonstrated the first application of Koopman theory and delay-coordinates embedding to linearize individual layers in a neural network. Adapting from extensive literature, we first reframed neural networks as a composition of different nonlinear maps, admitting neural activations as the states of a nonlinear dynamical system. We introduced layer scaling to augment the number of steps available for our states, building a trajectory matrix. Through analyzing the eigenvalues and right singular vectors of our trajectory matrices, we studied the use of delay-coordinates embedding as an observable function. For the Yin-Yang and MNIST datasets, we successfully replaced a fitted DMD model in trained MLPs, retaining significant performance and even visualizing the change in decision boundaries for the former dataset.

Considering our fresh perspective, our work is ripe with several questions to be addressed in future work. Most discernibly, developing further analyses to dissect the learned eigenmodes and eigenvalues of our fitted model would be insightful: can we further establish a link between what we observe about these components and how successful they will be in replacing a layer?

Of particular importance, given that we arrive at a linear operator, is the potential to explicitly understand and manipulate a trained layer. To highlight a few questions here, what do the varying success rates in hybridization across layers suggest about the role of each layer? How do we further develop our approach to decompose a network into its transformations, allowing for a mechanistic understanding? And, with vast literature in linear control to draw from, can we ``edit" trained models in the Koopman space? Finally, how do we build upon this framework to reliably accommodate any architectural choice (e.g. convolution, attention)?

Treating the layers of a neural networks as dynamical systems is a powerful framing, especially with the data-driven tendency of Koopman theory. We hope our work demonstrates the potential of this approach in advancing our understanding and control of neural networks.



\bibliography{main}

\clearpage

\appendix
\section{Yin-Yang MLP Training Details}
\subsection{Dataset} 
The Yin-Yang dataset \cite{kriener_yin-yang_2022} is a two-dimensional, publicly available classification task consisting of three categories: ``Yin", ``Yang", and ``Dot", allowing for easy visualization of the model's decision boundary. To begin with a problem which would prove straightforward for a neural network, we removed the ``Dot" class. Hence, in the modified dataset, each point in the dataset falls in either one of the two sides in the``Yin-Yang” symbol, and we leave the dots on both sides of the symbol empty.

\subsection{Architecture and Training} 
Table \ref{table:yinyang_arch} displays the architecture for the original MLP used to train on the Yin-Yang dataset. The MLP contains three hidden layers (IDs 1-3).
\begin{table}[h!]
\centering
\begin{tabular}{cccc}
\hline
\textbf{ID} & \textbf{Type} & \textbf{Input Dim} & \textbf{Output Dim} \\ \hline
0 & Linear + ReLU & 2 & 8 \\ \hline
1 & Linear + ReLU & 8 & 6 \\ \hline
2 & Linear + ReLU & 6 & 4 \\ \hline
3 & Linear + ReLU & 4 & 3 \\ \hline
4 & Linear & 3 & 2 \\ \hline
\end{tabular}
\caption{Architecture of the multi-layer perceptron (MLP) model for binary classification on the Yin-Yang dataset.}
\label{table:yinyang_arch}
\end{table}

We trained the original classifier for \ogYinYangEpochs epochs to an accuracy of \ogYinYangAccuracy, using the Adam with decoupled weight decay (AdamW) optimizer. We used a learning rate of \ogYinYangLR, \(\beta_1 = 0.9\), \(\beta_2 = 0.999\), and weight decay of \(1\mathrm{e}{-2}\). We train on \(2000\) randomly generated samples from the dataset, with a seed of \(42\) and a batch size of \(1000\) samples.

\subsection{Hyperparameter Tuning and Scaling}
To scale a hidden layer, we insert \YinYangScaledLayers additional Linear+ReLU layers directly before the layer we are interested in replacing. Before training the new layers, we searched for a learning rate and the AdamW betas, using Ray Tune (v2.34.0). Table 
\ref{table:yinyang_ray} presents the search space. 

\begin{table}[h!]
\centering
\begin{tabular}{cccccc}
\hline
\textbf{Hyperparameter} & \textbf{Search Space} \\ \hline
Learning rate & QLogUniform(\(1\mathrm{e}{-3}, 5\mathrm{e}{-3}, 1\mathrm{e}{-3}\)) \\ \hline
\(\beta_1\) & QLogUniform(\(0.2, 0.9, 1\mathrm{e}{-1}\)) \\ \hline
\(\beta_2\) & QLogUniform(\(0.5, 0.99, 1\mathrm{e}{-2}\)) \\ \hline
Weight decay & \(1\mathrm{e}{-3}\) \\ \hline
Batch size & \(512\) \\ \hline
\end{tabular}
\caption{Hyperparameter search space for the Yin-Yang scaled MLPs.}
\label{table:yinyang_ray}
\end{table}

We conducted this search for each hidden layer. Table \ref{table:yinyang_scaling_params} presents the final hyperparameters, along with the accuracy each scaled model achieved on the dataset.

\begin{table}[h!]
\centering
\begin{tabular}{cccccc}
\hline
\multicolumn{1}{l}{\textbf{ID}} & \multicolumn{1}{l}{\textbf{LR}} & \multicolumn{1}{l}{\textbf{\(\beta\) Values}} & \multicolumn{1}{l}{\textbf{\(\theta\) Decay}} & \multicolumn{1}{l}{\textbf{Batch}} & \multicolumn{1}{l}{\textbf{\begin{tabular}[c]{@{}l@{}}Test Acc.\\ (\%)\end{tabular}}} \\ \hline
1 & \multirow{3}{*}{2e-3} & \multirow{3}{*}{{[}0.8, 0.8{]}} & \multirow{3}{*}{1e-4} & \multirow{3}{*}{512} & 98.89 \\ \cline{1-1} \cline{6-6} 
2 &  &  &  &  & 98.88 \\ \cline{1-1} \cline{6-6} 
3 &  &  &  &  & 98.89 \\ \hline
\end{tabular}
\caption{Final hyperparameters and accuracy for scaled models trained on the Yin-Yang dataset, where the original model achieves an accuracy of 98.88\%.}
\label{table:yinyang_scaling_params}
\end{table}

\section{MNIST MLP Training Details}

\subsection{Dataset} 
We also conduct experiments with the MNIST digits dataset \cite{lecun_gradient-based_1998}, which is a 10-way digit classification task containing \(60,000\) training samples and \(10,000\) test samples.

\subsection{Architecture and Training} 
Table \ref{table:mnist_arch} shows the architecture for the MLP, with three hidden layers, used to train on the MNIST dataset.
\begin{table}[h!]
\centering
\begin{tabular}{cccc}
\hline
\textbf{ID} & \textbf{Type} & \textbf{Input Dim} & \textbf{Output Dim} \\ \hline
0 & Linear + ReLU & 784 & 256 \\ \hline
1 & Linear + ReLU & 256 & 128 \\ \hline
2 & Linear + ReLU & 128 & 64 \\ \hline
3 & Linear + ReLU & 64 & 32 \\ \hline
4 & Linear & 32 & 10 \\ \hline
\end{tabular}
\caption{Architecture of the multi-layer perceptron (MLP) model for 10-way classification on the MNIST dataset.}
\label{table:mnist_arch}
\end{table}

The MNIST classifer is trained for \ogEpochs epochs to an accuracy of \ogAccuracy. We use AdamW with a learning rate of \ogLR, \(\beta_1 = 0.9\), \(\beta_2 = 0.999\), a weight decay of \(1\mathrm{e}{-1}\), and a batch size of \(4096\) samples. In addition, we use a learning rate scheduler, which reduces the learning rate by a factor of \(0.5\) at a loss plateau with a patience of \(2\) epochs.

\subsection{Hyperparameter Tuning and Scaling}

\begin{table}[h!]
\centering
\begin{tabular}{cccccc}
\hline
\textbf{Hyperparameter} & \textbf{Search Space} \\ \hline
Learning rate & QLogUniform(\(1\mathrm{e}{-3}, 3\mathrm{e}{-3}, 1\mathrm{e}{-3}\)) \\ \hline
\(\beta_1\) & QLogUniform(\(0.6, 0.9, 1\mathrm{e}{-1}\)) \\ \hline
\(\beta_2\) & QLogUniform(\(0.7, 0.99, 1\mathrm{e}{-2}\)) \\ \hline
Weight decay & \(1\mathrm{e}{-2}\) \\ \hline
Batch size & \(4096\) \\ \hline
\end{tabular}
\caption{Hyperparameter search space for the MNIST scaled MLPs.}
\label{table:mnist_ray}
\end{table}

For scaling, once again, we insert \YinYangScaledLayers additional Linear+ReLU layers and conduct a hyperparameter search before training. Table  \ref{table:mnist_ray} presents the search space and Table \ref{table:mnist_scaling_params} presents the final hyperparameters and test accuracies. 

\begin{table}[ht!]
\centering
\begin{tabular}{cccccc}
\hline
\multicolumn{1}{l}{\textbf{ID}} & \multicolumn{1}{l}{\textbf{LR}} & \multicolumn{1}{l}{\textbf{\(\beta\) Values}} & \multicolumn{1}{l}{\textbf{\(\theta\) Decay}} & \multicolumn{1}{l}{\textbf{Batch}} & \multicolumn{1}{l}{\textbf{\begin{tabular}[c]{@{}l@{}}Test Acc.\\ (\%)\end{tabular}}} \\ \hline
1 & \multirow{2}{*}{2e-3} & {[}0.7, 0.7{]} & \multirow{3}{*}{1e-2} & \multirow{3}{*}{4096} & 96.63 \\ \cline{1-1} \cline{3-3} \cline{6-6} 
2 &  & {[}0.9, 0.85{]} &  &  & 96.71 \\ \cline{1-3} \cline{6-6} 
3 & 3e-3 & {[}0.9, 0.99{]} &  &  & 96.82 \\ \hline
\end{tabular}
\caption{Final hyperparameters and accuracy for scaled models trained on the Yin-Yang dataset, where the original model achieves an accuracy of 98.88\%.}
\label{table:mnist_scaling_params}
\end{table}

\makeatletter
\setlength{\@fptop}{0pt}
\makeatother



\end{document}